\documentclass[runningheads]{llncs}
\usepackage[T1]{fontenc}
\usepackage{graphicx}
\usepackage{booktabs}
\usepackage{amsmath}
\usepackage{amssymb}
\usepackage{xcolor}
\begin{document}
\title{From 50\% to Mastery in 3 Days: A Low-Resource SOP for Localizing Graduate-Level AI Tutors via Shadow-RAG}
\titlerunning{A Low-Resource SOP for Localizing AI Tutors}
\author{Zonglin Yang\inst{1,2} \and
J.-H. Xie\inst{1} \and
Lining Zhang\inst{3} \and
Jiyou Jia\inst{3}\thanks{Corresponding author: jjy@pku.edu.cn} \and
Zhi-X. Chen\inst{1,4}\thanks{Corresponding author: chenzhi@pku.edu.cn}}

\authorrunning{Yang et al.}

\institute{
School of Mechanics and Engineering Science, Peking University, Beijing, 100871, China
\and
Readraft Intelligent Technology Co., Ltd., China
\and
Graduate School of Education, Peking University, Beijing, 100871, China
\and
AI for Science Institute, Beijing, 100080, China
}

\maketitle
\begin{abstract}
Deploying high-fidelity AI tutors in schools is often blocked by the Resource Curse---the need for expensive cloud GPUs and massive data engineering. In this practitioner report, we present a replicable Standard Operating Procedure that breaks this barrier. Using a Vision-Language Model data cleaning strategy and a novel Shadow-RAG architecture, we localized a graduate-level Applied Mathematics tutor using only 3 person-days of non-expert labor and open-weights 32B models deployable on a single consumer-grade GPU. Our pilot study on a full graduate-level final exam reveals a striking emergence phenomenon: while both zero-shot baselines and standard retrieval stagnate around 50-60\% accuracy across model generations, the Shadow Agent, which provides structured reasoning guidance, triggers a massive capability surge in newer 32B models, boosting performance from 74\% (Naive RAG) to mastery level (90\%). In contrast, older models see only modest gains ($\sim$10\%). This suggests that such guidance is the key to unlocking the latent power of modern small language models. This work offers a cost-effective, scientifically grounded blueprint for ubiquitous AI education.

\keywords{Low-Resource AI \and Educational SOP \and Shadow RAG \and Local Deployment}
\end{abstract}
\section{Introduction: Breaking the Resource Curse}

The promise of ``AI for All'' in education is currently stalled by a ``Resource Curse''~\cite{campbell2025}.
This curse manifests as a confluence of three interconnected barriers.
First, institutions face a compute-privacy dilemma: while proprietary cloud models (e.g., GPT-4o) possess the necessary reasoning capabilities, they are often inaccessible due to strict data privacy regulations for educational institutions~\cite{moe2021,pipl2021,cac2022} and prohibitive cluster costs.
Conversely, privacy-safe local models deployable on consumer-grade GPUs often lack the latent intelligence required for graduate-level STEM reasoning~\cite{wei2022}.

Second, a data bottleneck traps valuable pedagogical knowledge in unstructured formats---such as handwritten blackboard notes and lecture screenshots---that defy standard digitization pipelines.

Crucially, we confront a third barrier: the failure of naive retrieval in low-resource settings.
We find that standard Retrieval-Augmented Generation (RAG)~\cite{lewis2020} is insufficient for complex problem-solving; merely retrieving a mathematical definition is futile if the weaker local model cannot logically determine the constraints and methodology required to apply it.

In this practitioner report, we describe a localized implementation that bypasses these barriers.
We ask: \emph{Can a small team with limited resources build a state-of-the-art (SOTA) graduate tutor?}
We answer with a replicable Standard Operating Procedure (SOP) designed to dismantle this curse.
By combining a Vision-Language Model (VLM) data cleaning workflow with a novel Shadow-RAG architecture~\cite{liu2025}, we demonstrate how to empower a standard 32B open-weights model to achieve mastery-level performance.
This approach enabled us to convert a semester of raw Applied Mathematics notes into a reliable AI tutor in just 3 days using non-expert labor.
Unlike many educational tools that prioritize glossy UI/UX~\cite{campbell2025}, we argue that mathematical reliability must precede interactive design~\cite{cac2023}.
Therefore, this work focuses squarely on the backend reasoning engine, offering a cost-effective, scientifically grounded blueprint for ubiquitous AI education.

% [INSERT FIGURE 1 HERE]
\begin{figure}[t]
\centering
\includegraphics[width=1.0\textwidth]{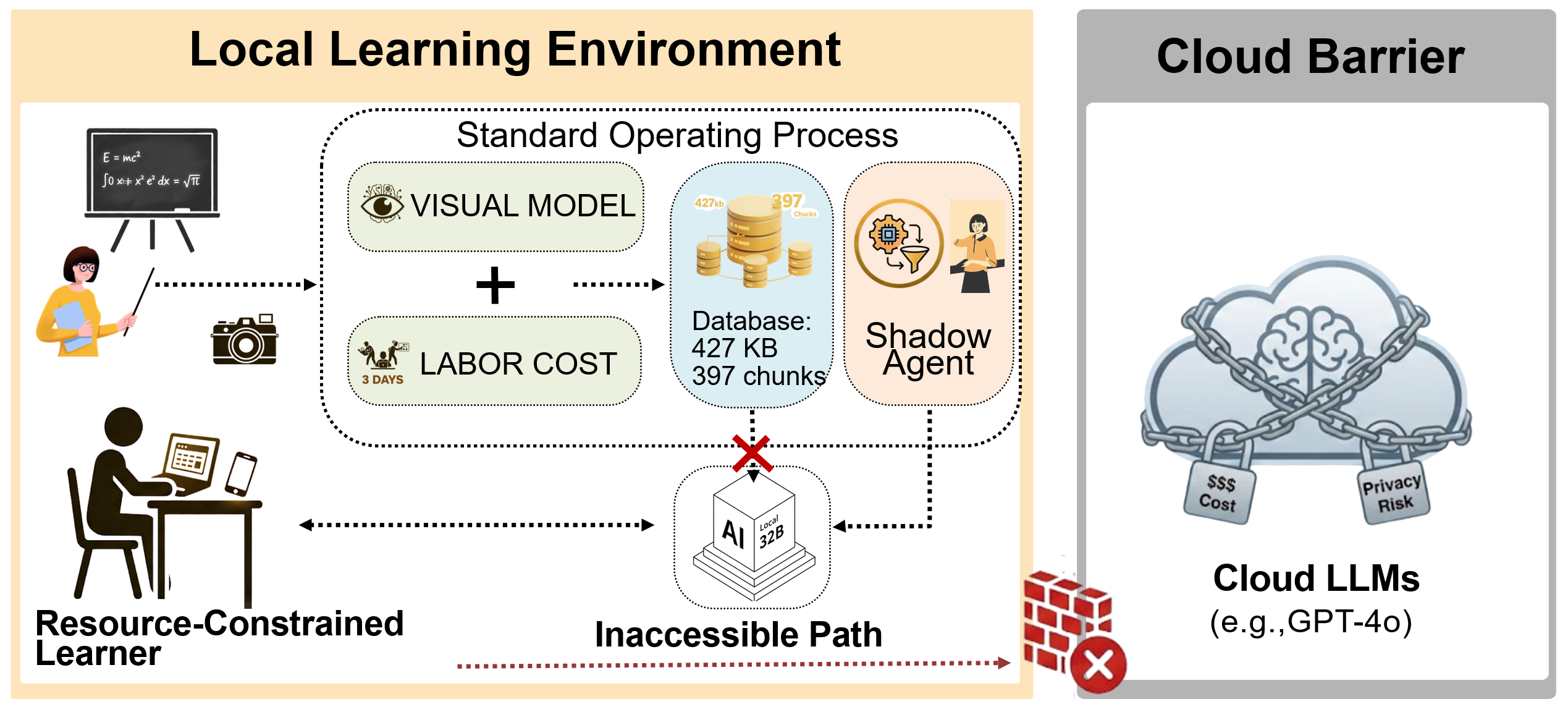}
\caption{\textbf{Breaking the Barriers.} The "Resource Curse" (Right) blocks access to Cloud AI. Our SOP (Left to Center) utilizes VLM-assisted cleaning (3 Days) and a Shadow Agent to empower a local 32B model, effectively bypassing the cloud wall.} \label{fig1}
\end{figure}

\section{Description of the Implementation}

Our implementation targets a graduate-level "Advanced Applied Mathematics" course at a leading research university in China (consequently, all notes and problem sets are in Chinese). We selected this context because it is highly representative: across global higher education—even at elite institutions—advanced STEM instruction heavily relies on unstructured blackboard teaching, presenting a universal data bottleneck.

The participating learners are STEM graduate students engaged in rigorous theoretical studies. For these learners, graduate-level problems are exceptionally difficult, and reliable reference materials are notoriously scarce. Since professors lack the time for daily, personalized tutoring, our system is designed to ultimately act as a constant reasoning teammate for the students. While the ultimate objective is to provide a 24/7 localized AI Teaching Assistant (TA), we identified a critical prerequisite: current privacy-safe local models frequently hallucinate and fail to solve these advanced problems correctly in the first place. Therefore, our primary educational goal in this phase is to establish backend mathematical reliability. We argue that an AI must be capable of generating correct, rigorous solutions autonomously before any meaningful interactive tutoring features can be evaluated. Consequently, our implementation focuses entirely on building a highly accurate reasoning engine driven by open-weights models, operating within strict campus data privacy policies.

Our Standard Operating Procedure (SOP) to achieve this consists of two phases:

\subsection{Phase 1: Zero-Cost Data Engineering via Human-AI Synergy}

Our data acquisition strategy targeted the messy reality of classroom instruction: 45 class hours of pure blackboard teaching with no accompanying slides. Manually transcribing this material would require hundreds of hours—time that professors and teaching assistants simply do not have. To overcome this, our SOP treats AI not just as a tool, but as a co-creation teammate for educators. We collected approximately 150 raw screenshots from lecture videos and student photos, featuring typical noise such as glare, cursive handwriting, and non-linear layouts. To process this unstructured raw material, we employed a state-of-the-art Vision-Language Model (VLM, Gemini 3 Pro) as a semantic transcriber. Unlike traditional OCR, we prompted the model to "organize the content into logically rigorous lecture notes with detailed derivations," thereby enforcing mathematical coherence during the digitization process.

This approach demonstrates a highly practical human-AI synergy. The human role was strictly limited to syntax verification and acting as a logical gatekeeper. Annotators focused on segmenting content, removing AI-generated artifacts, and structuring the output, without requiring deep domain expertise. Through this collaborative pipeline, we transformed a labor-intensive chore into an efficient process: producing a final corpus of 103 Markdown files (approximately 428KB), which were subsequently segmented into 397 logical chunks using a double-newline strategy with an 800-character merging threshold, in just 3 person-days.
\subsection{Phase 2: The Shadow-RAG Architecture}

% [INSERT FIGURE 2 HERE]
\begin{figure}[hbt]
\centering
\includegraphics[width=0.8\textwidth]{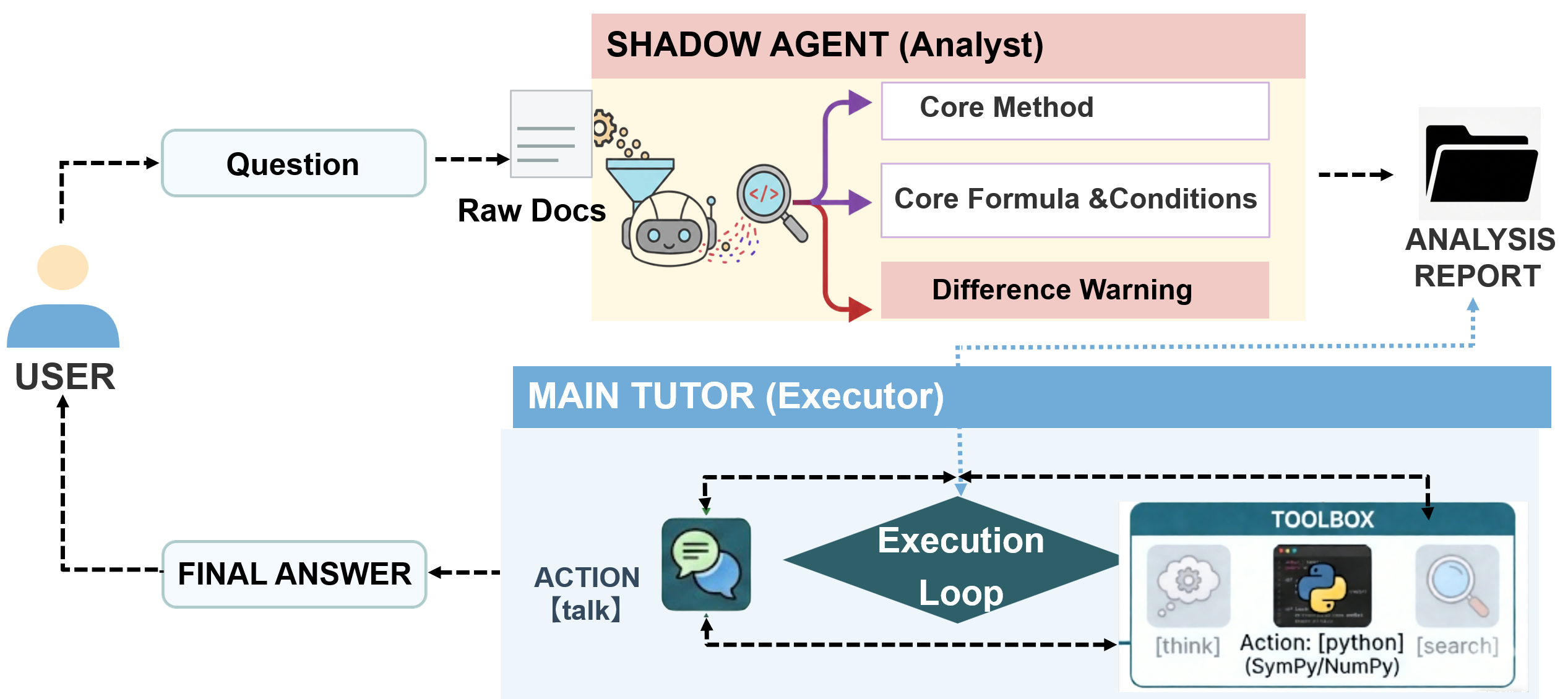}
\caption{\textbf{Shadow-RAG Logic Flow.} The Shadow Agent intercepts raw chunks and distills them into methodological guidance. The Main Tutor then acts on this guidance to choose between Direct answering or Tool usage.} \label{fig2}
\end{figure}

To run effectively on a mid-sized local model (Qwen2.5/3-32B), we moved away from "Naive RAG" (Retrieval-Augmented Generation)—the standard practice of simply retrieving relevant text chunks and feeding them to the model—to a dual-agent architecture that mirrors the analytical process of open-book learning.

The system relies on a \textit{Shadow Agent} acting as a background "Intelligence Analyst." Governed by a system prompt that enforces strict scrutiny, this agent never interacts directly with the user. Instead, it reads the raw, retrieved mathematical chunks and outputs a structured analytical report consisting of: (1) Core Method Extraction, (2) Formulaic Conditions, and crucially, (3) A "Difference Warning" (e.g., "Warning: The stationary point is at the boundary, not the interior, leading to coefficient variations!"). We term this the "Shadow" Agent not only for its background role in denoise filtering, but also as a metaphor for its mission: to illuminate the "shadows" of the digital divide by enabling powerful reasoning on constrained hardware.

The \textit{Main Tutor} acts as the "Strategic Executor." Instead of reading raw RAG text—which often causes local models to mechanically apply formulas without comprehension—it only reads the Shadow Agent's distilled report. Based on the discrepancies identified by the Shadow Agent between the query and the retrieved notes, the Main Tutor autonomously determines the optimal resolution path: either performing direct, intuitive reasoning through a \texttt{[talk]} action when the problem aligns with established knowledge, or triggering a programmatic derivation via a \texttt{[python]} action (SymPy/NumPy) to re-verify solutions from first principles when significant variations are detected.

\subsubsection{Pedagogical Insight}
The educational insight behind this architecture is that advanced mathematics relies heavily on \textit{conditional applicability}—a nuance Naive RAG systematically fails to grasp. Standard RAG models tend to act as mere search tools, blindly copying retrieved formulas. By mandating a Difference Warning, our Shadow Agent simulates human pedagogical scrutiny—verifying logical boundaries before the Main Tutor acts. This internal AI synergy effectively replaces unreliable generative tools with a trustworthy educational teammate.

\section{Reflection of Challenges and Opportunities}

Rather than participating in standard scientific benchmarking on public datasets, we sought a practical, domain-specific acceptance metric. We posit that if a rigorous graduate-level final exam is the gold standard for assessing human mastery, it logically serves as a valid test for an AI's reasoning capabilities within this vertical domain.
\subsection{Experimental Definitions and Protocol}
To rigorously evaluate our system against this metric, we designed a controlled pilot study simulating local deployment using API endpoints for two representative local models (Qwen2.5-32B-Instruct and Qwen3-32B). Our test set comprised a full final exam ($N=5$ major problems). To protect course confidentiality, we applied a homomorphic rewrite to these questions—altering variables and contexts while strictly preserving the original mathematical concepts and difficulty level.

The protocol evaluated both models across five distinct operational configurations. To account for generation variance, each configuration was tested five times per question, yielding 250 total inferences. The outputs were evaluated by a strong LLM judge (deepseek-ai/DeepSeek-V3), which inherently possesses the capability to independently solve these advanced problems correctly. To eliminate arbitrary holistic scoring, the evaluation strictly followed a customized, step-by-step grading rubric.

To isolate the impact of our proposed architecture, we structured the evaluation as an ablation study. We first established a performance floor using a Baseline (Zero-shot) and a Naive RAG setup, representing the standard retrieval approach without agentic guidance. We then evaluated three distinct engineering strategies of the Shadow methodology:

\begin{description}
    \item[Shadow (Full/Dynamic)] The model acts as an agent that autonomously chooses between text reasoning and Python tools.
    \item[Shadow (No Code)] The model is restricted to pure text-based reasoning to test the guidance alone.
    \item[Shadow (Forced Tools)] The model is mandated to perform programmatic calculation at every step.
\end{description}

This comparative design allowed us to investigate the optimal implementation strategy for the Shadow methodology, analyzing the interplay between programmatic execution and theoretical reasoning across varying base model capabilities. Table \ref{tab:configs} summarizes these configurations.

\begin{table}[htbp]
\centering
\caption{Experimental Configurations in the Ablation Study.}
\label{tab:configs}

\begin{tabular*}{\textwidth}{@{\extracolsep{\fill}}lcccc@{}}
\toprule
\textbf{Configuration} & \textbf{RAG} & \textbf{Shadow} & \textbf{Reason} & \textbf{Code} \\ \midrule
Baseline (Zero-shot) & $\times$ & $\times$ & $\checkmark$ & $\times$ \\
Naive RAG & $\checkmark$ & $\times$ & $\checkmark$ & $\times$ \\
Shadow (Full/Dynamic) & $\checkmark$ & $\checkmark$ & $\checkmark$ & $\checkmark$ \\
Shadow (No Code) & $\checkmark$ & $\checkmark$ & $\checkmark$ & $\times$ \\
Shadow (Forced Tools) & $\checkmark$ & $\checkmark$ & $\times$ & $\checkmark$ \\ \bottomrule
\end{tabular*}
\end{table}

\subsection{Evaluation Outcomes}

We report the quantitative outcomes of the pilot study in Table 2 below.

\begin{table}[htbp]
\centering
\small
\caption{Pilot study results: accuracy across configurations and model generations}
\label{tab:pilot_results}
\begin{tabular*}{\textwidth}{@{\extracolsep{\fill}} lcc}
\toprule
Configuration             & Qwen2.5-32B Acc. (\%) & Qwen3-32B Acc. (\%) \\
\midrule
Baseline (Zero-shot)      & 47 & 67 \\
Naive RAG                 & 56 & 74 \\
Shadow (Full/Dynamic)     & 65 & 85 \\
Shadow (No Code)          & 50 & 85 \\
Shadow (Forced Tools)       & 57 & 90 \\
\bottomrule
\end{tabular*}
\end{table}

For the stronger model, Qwen3-32B, the Baseline configuration achieved an average accuracy of 67\%, and Naive RAG raised this to 74\%. However, all three Shadow variants moved the model into a significantly higher performance regime. Notably, Shadow-Dynamic and Shadow-NoCode both reached 85\%, while the constraint-heavy Shadow-Forced peaked at 90\%. Thus, once the Shadow architecture is enabled, performance clusters in the mid-80\% to 90\% range, establishing a mastery plateau well above standard retrieval methods.

For the weaker model, Qwen2.5-32B, the pattern is noticeably different. The Baseline accuracy sits at 47\%, with Naive RAG increasing it modestly to 56\%. While the Shadow variants yield further gains, the overall scores remain confined to the mid-50\% to mid-60\% band. Interestingly, unlike the stronger model, Qwen2.5 performed best with the flexible Shadow-Dynamic configuration (65\%), outperforming the rigid Shadow-Forced setup (57\%). This suggests that weaker models struggle with strict constraints and benefit from the autonomy to choose their solving path.

\begin{figure}[t]
\centering
\includegraphics[width=1.0\textwidth]{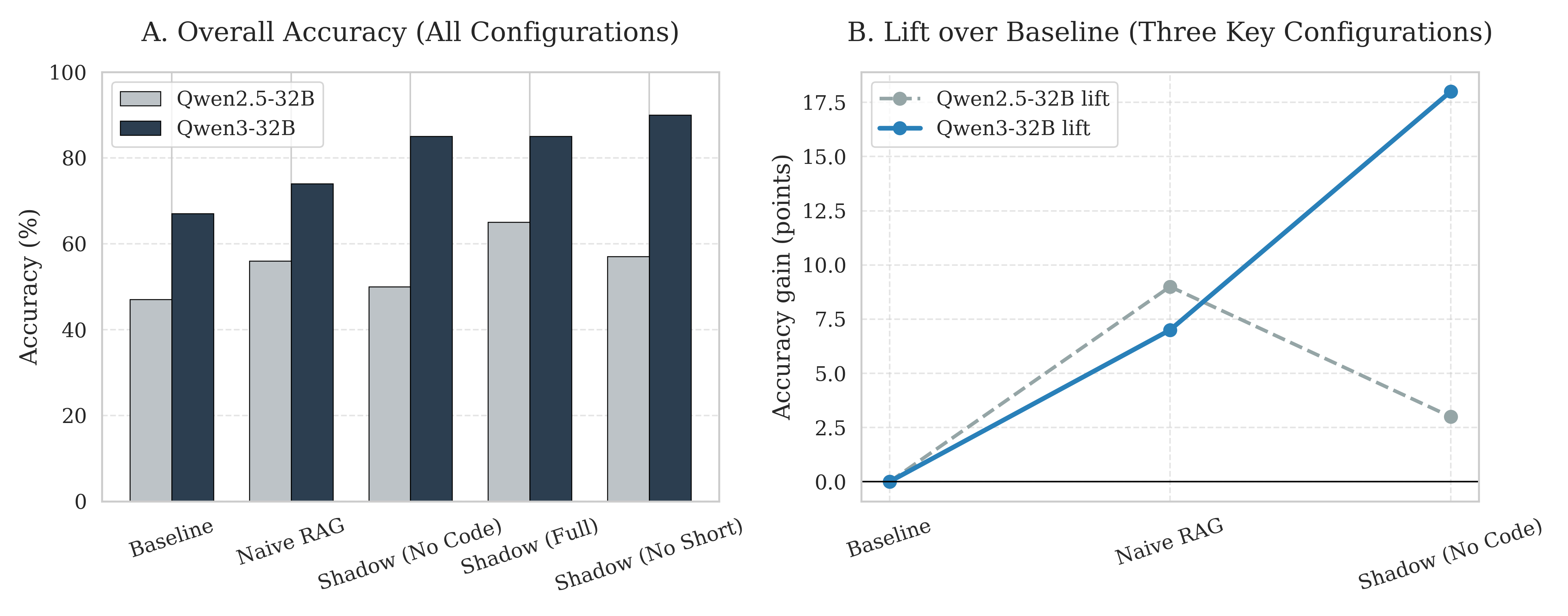}
\caption{\textbf{Performance and Gains.} (Left) Absolute accuracy scores showing the Qwen3 plateau. (Right) Improvement over baseline, highlighting the non-linear "Emergence" in Qwen3 with Shadow methodology.} \label{fig3}
\end{figure}

Figure \ref{fig3} (left) visualizes these scores. Qwen3 shows a clear separation between non-Shadow and Shadow settings, forming a high-performance plateau. Qwen2.5 exhibits a downward shift, with a preference for dynamic tool use over forced execution.

To highlight this difference, Figure \ref{fig3} (right) replots gains over Baseline. For Qwen2.5, Naive RAG provides a $\sim$9\% gain, and Shadow-NoCode adds negligible improvement (50\% vs 47\%). In stark contrast, for Qwen3, switching to Shadow-NoCode produces a massive jump from Naive RAG's 74\% to 85\%. This generation-dependent response highlights an intriguing "Emergence" phenomenon: the stronger model exhibits a non-linear capability surge when architectural scaffolding is introduced, whereas the weaker model shows only linear, incremental gains.

\subsection{Key Reflection: The Role of Methodology}

The quantitative divergence described above leads to a critical insight: while engineering optimizations (like tool use) matter, the methodological scaffolding provided by the Shadow Agent is the primary driver of logical reliability. By establishing a rigorous reasoning framework, the Shadow architecture propelled Qwen3 from 74\% to a mastery level of 90\% (Shadow-Forced). Even without code execution, the architecture alone achieved 85\%, proving that the primary driver was reasoning orientation, which was then further amplified by computational tools.

To elucidate the mechanism behind this gain, we analyze a representative case involving the asymptotic expansion of an oscillatory integral over a finite range $[0, 1]$.

\subsubsection*{The Retrieval Trap}
Standard RAG retrieved a stationary phase formula derived for an \textit{infinite} domain $(-\infty, +\infty)$. The baseline model, failing to check the validity scope, blindly applied this generic formula and hallucinated an incorrect coefficient.

\subsubsection*{The Shadow Correction}
In contrast, the Shadow Agent performed a critical boundary audit. It detected that the stationary point coincided with the integration endpoint, triggering a specific warning:
\begin{quote}
    \textit{``Warning: The stationary point is at the endpoint... Do not apply the full-range formula. Apply the half-range Fresnel integral.''}
\end{quote}
Guided by this qualitative logic, the system correctly adjusted the result by a factor of $1/2$. This demonstrates the Shadow Agent's core value: it simulates an expert teacher effectively validating boundary conditions---a capability that prevents the contextual errors common in naive retrieval.

Ultimately, this study presents a pragmatic commercial and educational opportunity by addressing the dual bottlenecks of hardware and data costs. Historically, deploying high-fidelity STEM tutors required expensive cloud clusters and expert-level data annotation. Our results demonstrate a cost-effective alternative: by combining open-weight Small Language Models (SLMs) with a streamlined, low-labor data cleaning SOP and the logic-gating Shadow architecture, we achieved mastery-level performance using only consumer-grade hardware. This approach drastically lowers the barrier to entry, suggesting that scalable, domain-specific AI tutors can be developed and deployed locally by practitioners with limited budgets and non-expert teams, making high-quality AI education more accessible.

\subsection{Challenges and Areas for Improvement}

While the Shadow-RAG architecture demonstrates high efficacy, we confront a significant trade-off between reasoning depth and interaction fluidity that hinders immediate scalable deployment. The dual-agent reasoning process necessitates a 10-fold increase in token consumption to secure this performance gain; thus, despite equipping the system with a conversational \texttt{[talk]} tool to mitigate user anxiety, the resulting latency remains a critical challenge for the product form, straining the ideal of a fluid, real-time educational "teammate." Furthermore, the system exhibits significant sensitivity to the underlying model. Our ablation results demonstrate that while the Shadow Agent serves as the primary driver of performance improvement, the specific protocols for tool selection require case-by-case calibration to align with the varying latent capabilities of different base models.

\section{Future Steps}

Having established the system's baseline reliability, our future work prioritizes three strategic dimensions. First, we aim to validate universality by initially extending the Shadow-RAG architecture to adjacent STEM domains to verify the transferability of our logic-gating mechanism. Crucially, to facilitate the transition from a static "tool" to a real-time "teammate," we will prioritize identifying a viable product form for practical deployment to gather authentic student feedback and empirically validate the system's pedagogical efficacy in real-world settings. Finally, we highlight the intriguing "Emergence" phenomenon observed in our study, where structured guidance triggered a non-linear capability surge in the Qwen3-32B model (74\% to 90\%). We invite the research community to investigate this mechanism, as we believe it is the key to optimizing the performance of Small Language Models in resource-constrained environments.

\end{document}